\documentclass[letterpaper]{article} 
\usepackage{aaai2026}  
\usepackage{times}  
\usepackage{helvet}  
\usepackage{courier}  
\usepackage[hyphens]{url}  
\usepackage{graphicx} 
\usepackage{natbib}  
\usepackage{caption} 
\usepackage{algorithm}
\usepackage{algorithmic}
\usepackage{amsmath, amssymb, amsthm}
\usepackage{amsfonts}       
\usepackage{newfloat}
\usepackage{listings}
\DeclareCaptionStyle{ruled}{labelfont=normalfont,labelsep=colon,strut=off} 
\floatstyle{ruled}
\newfloat{listing}{tb}{lst}{}
\floatname{listing}{Listing}
\title{Grounding vs. Compositionality: On the Non-Complementarity of Reasoning in Neuro-Symbolic Systems}
\author {
    Mahnoor Shahid\textsuperscript{\rm 1,2},
    Hannes Rothe\textsuperscript{\rm 1,3}
}
\affiliations {
    \textsuperscript{\rm 1}Universität Duisburg Essen, Germany\\
    \textsuperscript{\rm 2}mahnoor.shahid@uni-due.de, \textsuperscript{\rm 3}hannes.rothe@uni-due.de
}

\usepackage{bibentry}

\begin{document}

\maketitle

\begin{abstract}
Compositional generalization remains a foundational weakness of modern neural networks, limiting their robustness and applicability in domains requiring out-of-distribution reasoning. A central, yet unverified, assumption in neuro-symbolic AI is that compositional reasoning will emerge as a byproduct of successful symbol grounding. This work presents the first systematic empirical analysis to challenge this assumption by disentangling the contributions of grounding and reasoning. 
To operationalize this investigation, we introduce the Iterative Logic Tensor Network ($i$LTN), a fully differentiable architecture designed for multi-step deduction. Using a formal taxonomy of generalization---probing for novel entities, unseen relations, and complex rule compositions---we demonstrate that a model trained solely on a grounding objective fails to generalize. In contrast, our full $i$LTN, trained jointly on perceptual grounding and multi-step reasoning, achieves high zero-shot accuracy across all tasks. Our findings provide conclusive evidence that symbol grounding, while necessary, is insufficient for generalization, establishing that reasoning is not an emergent property but a distinct capability that requires an explicit learning objective.
\end{abstract}

\begin{links}
    \link{Code}{https://github.com/Place-Beyond-Bytes/grounding_vs_compositionality.git}
\end{links}

\section{Introduction}\label{sec:introduction}

Compositional Generalization is the ability to understand and generate novel combinations from known components~\cite{frankland2020concepts,lin2023survey,hupkes2020compositionality}. Humans effortlessly understand novel combinations of familiar concepts~\cite{lake2017building}. However, this capacity for compositionality is a foundational weakness of modern neural networks \cite{fodor1988connectionism}. 
Even state-of-the-art models still exhibit profound brittleness when confronted with such compositional shifts~\cite{woydt2025fodor}. This gap hinders their application in domains requiring robust, out-of-distribution reasoning~\cite{lake2023human}. Neuro-symbolic AI has emerged as a promising paradigm to address this brittleness by integrating the perceptual strengths of neural networks with the structured reasoning capabilities of symbolic logic~\cite{bhuyan2024neuro,feldstein2024mapping,sheth2023neurosymbolic}. The objective is to build models that can ground high-dimensional perceptual input into a symbolic vocabulary and subsequently perform structured reasoning over the resulting representations.

Within this field, there is often an implicit assumption that compositionality will emerge naturally or intrinsically by default from symbolic components once grounding is achieved~\cite{garcez2023neurosymbolic,marcus2003algebraic,pavlick2023symbols,wu2024a}. The intuition is that if a model can reliably map pixels to symbols (e.g., `3', `blue', `cube'), it will inherently be able to use those symbols in new logical deductions \cite{andreas2016neural,mao2019neuro}. However, this hypothesis remains largely unverified.
Empirical studies tend to focus on either symbol grounding in rich perceptual settings but without systematic generalization tests \cite{antol2015vqa,topan2021techniques,li2024softened,umili2023grounding}, or on compositional generalization in purely symbolic domains that abstract away perception \cite{conklin2021meta,lake2018generalization,lightman2023let}, rarely evaluating them in a unified, controlled setting. This creates a critical gap in our understanding of whether these two capabilities are intrinsically linked or represent distinct, non-complementary challenges.

This gap in empirical evidence is what this research aims to address. Our study thereby asks: \textit{Is compositional generalization an emergent property of symbol grounding, or does it require a distinct learning objective?} To empirically investigate the link between SG and CG, we require an architecture capable of multi-step, differentiable reasoning. We identify a class of iterative, state-refining architectures, more suitable for this purpose. Therefore, we introduce the Iterative Logic Tensor Network ($i$LTN), which instantiates this paradigm within a differentiable logic framework, allowing us to directly manipulate and evaluate logical rule composition. To isolate the contributions of both grounding and reasoning, we conduct a controlled study comparing three models: (1) a baseline trained purely on a symbol grounding objective; (2) our proposed $i$LTN, trained with objectives for both grounding and multi-step reasoning; and (3) an ablated version of $i$LTN trained only with the reasoning objective, to test if grounding is a necessary component. We evaluate all three models on a rigorous, formal taxonomy of generalization designed to probe three distinct axes of compositionality: generalization to (1) novel entities, (2) unseen relations, and (3) multi-step rule compositions.

Our findings reveal a critical dependency: while symbol grounding is a necessary foundation, it alone is insufficient for achieving robust compositional generalization. Our full $i$LTN model, which combines grounding with an explicit reasoning objective, significantly outperforms both a grounding-only baseline and an ablated version of $i$LTN without a grounding mechanism, establishing that both components are essential. Our main contributions are 3-fold:
\begin{enumerate}
    \item To the best of our knowledge, we provide the first empirical analysis to systematically investigate the distinct contributions of symbol grounding and compositional reasoning, providing quantitative evidence that symbol grounding is necessary but not sufficient for compositional generalization.
    \item We introduce the Iterative Logic Tensor Network ($i$LTN), a fully differentiable architecture that operationalizes multi-step logical deduction as an iterative refinement process, enabling generalization to complex reasoning chains.
    \item We provide strong evidence for the superiority of a joint learning approach. Our targeted ablation studies show that a model trained end-to-end on both grounding and reasoning significantly outperforms specialized models trained on grounding alone, or a model trained on reasoning alone.
\end{enumerate}

\section{Related Work} \label{sec:related_work} 
Our research is positioned at the intersection of two distinct but related fields: symbol grounding and compositional generalization. While many NeSy systems aim to integrate these capabilities, we argue that they have evolved in relative isolation, leaving the precise relationship between grounding and compositional reasoning largely unexplored. 

\subsection{Perceptual Grounding in Neurosymbolic Learning}
The benefits of symbolic grounding for domain generalization and data efficiency have been shown in previous neurosymbolic learning research \cite{andreas2016learning,he2016deep,zhan2021unsupervised,shah2020learning,mao2019program,Hsu_2023_CVPR}.

The synergy between neural and symbolic components can be organized into three main groups, based on how they integrate neural and symbolic representations: (1) using pre-computed symbols to augment perceptual inputs for a downstream task \cite{mao2019program,ellis2018learning,andreas2016learning,Hsu_2023_CVPR}, (2) deriving useful symbolic abstraction directly from perceptual data \cite{tang2023perception}, and (3) jointly learning both neural and symbolic encodings for prediction \cite{zhan2021unsupervised,sehgal2024neurosymbolic}.

Our work falls into this third category, building on the well-established principle that symbolic grounding facilitates domain generalization and improves data efficiency. However, a key challenge remains within this paradigm. The prominent approach by Zhan et al. (2021), for instance, also jointly learns representations but employs a program synthesizer to search over a fixed Domain-Specific Language (DSL). This reliance on program synthesis can introduce significant bottlenecks in terms of scalability and expressive power, a limitation our work aims to address by considering a fully differentiable, gradient-based framework.

\subsection{Compositional Generalization}
The failure of neural networks to generalize compositionally~\cite{fodor1988connectionism} has been empirically confirmed by a series of diagnostic benchmarks. The SCAN dataset~\cite{lake2018generalization} revealed the failures of sequence-to-sequence models on novel combinations of commands. Similarly, datasets like CLEVR~\cite{johnson2017clevr}, gSCAN~\cite{ruis2020benchmark}, and COGS~\cite{kim2020cogs} further demonstrated that even large models rely on spurious correlations rather than genuine compositional reasoning. 
In response, proposed solutions have primarily fallen into two families: 
(1) Neural approaches enforcing compositionality by design (Neural Module Networks (NMNs)~\cite{andreas2016neural} and their successors like the MAC network~\cite{hudson2018compositional} dynamically assemble specialized neural modules into a computational graph to solve a task. While effective, these methods often rely on a fixed set of modules and struggle to generalize to problems requiring longer or more complex reasoning chains.  (2) Neuro-symbolic (NeSy) approaches take a different route, translating perceptual input into symbolic representations that are then processed by a logical reasoner. Models like the NS-CL~\cite{mao2019neuro} and NS-VQA~\cite{yi2018neural} have demonstrated strong compositional abilities, but their reliance on non-differentiable symbolic backbones can hinder end-to-end learning.

Our work addresses these limitations by building on differentiable logic, using Logic Tensor Networks (LTNs)~\cite{serafini2016logic}. Standard LTNs perform one-shot satisfiability checks and cannot model sequential deduction required for multi-hop reasoning. 
To overcome this, the iterative refinement process of our proposed $i$LTN is inspired by the recurrent, multi-step reasoning paradigm established by architectures like the MAC network. However, where the MAC network implements its reasoning steps using a fully neural, attention-based architecture, the $i$LTN instantiates this iterative process within a differentiable first-order logic framework. This distinction is critical for our central research goal: it provides a transparent and controllable symbolic backend that enables the model to systematically compose explicit logical axioms. This capability is essential for empirically testing our taxonomy of generalization—particularly for relational and rule composition—and analyzing the interplay between grounding and reasoning in a controlled manner.

\section{Problem Definition}\label{sec:problem_statement}

This study incorporates three distinct and increasingly challenging forms of compositional generalization (CompGen) within a unified neuro-symbolic framework. We formalize our investigation as a series of Neuro-Symbolic Constraint Satisfaction Problems (NS-CSPs) designed to test a model's ability to generalize beyond its training distribution. Our central hypothesis is that robust generalization requires more than simple symbol grounding; it requires explicit architectural support for reasoning about novel entities, relations, and rule compositions.

\subsection{Visual Logic Puzzle Domain \& Symbol Grounding}
Our task domain consists of visual logic puzzles where the underlying state is decomposable and governed by a symbolic knowledge base.

\textbf{State Space:} The model observes an initial puzzle configuration as an image $I \in \mathcal{S}_{\text{pixel}}$. We assume this high-dimensional state is grounded in an object-centric latent space $\mathcal{Z} = \mathcal{Z}_{1,1} \times \dots \times \mathcal{Z}_{N,N}$, where each $\mathcal{Z}_{r,c}$ is the latent representation for the cell at $(r,c)$. A neural perception model learns this initial mapping function, $G_{\theta}: \mathcal{S}_{\text{pixel}} \rightarrow \mathcal{Z}$.

\textbf{Symbol Grounding:} A core challenge is connecting these latent representations to their discrete, symbolic meaning. We define a symbolic vocabulary $\mathcal{V}$ containing all unique symbols in the domain (e.g., for Sudoku, $\mathcal{V} = \{1, 2, \dots, 9, \text{blank}\}$). Then, we learn a projection function $\Pi: \mathcal{Z}_{r,c} \rightarrow \mathcal{V}$ that correctly maps the latent vector of a cell to its corresponding discrete symbol. Our investigation begins by testing whether successfully solving for $\Pi$ is sufficient to achieve downstream compositional generalization.

\textbf{Symbolic Knowledge Base ($\mathcal{K}$):} The puzzle's logic is defined by a knowledge base $\mathcal{K}$, a set of first-order logic (FOL) axioms that operate on the grounded symbols from $\mathcal{V}$. We define a hierarchy of knowledge bases, where $\mathcal{K}_{\text{easy}}$ contains the fundamental rules, $\mathcal{K}_{\text{moderate}}$ extends this set with axioms for intermediate strategies, and $\mathcal{K}_{\text{hard}}$  further adds axioms for complex, multi-step strategies. This creates a nested hierarchy where $\mathcal{K}_{\text{easy}} \subset \mathcal{K}_{\text{moderate}} \subset \mathcal{K}_{\text{hard}}$. 

\subsection{A Taxonomy of Compositional Generalization}
We define CompGen as a distributional shift between training ($\mathcal{D}_{\text{train}}$) and evaluation ($\mathcal{D}_{\text{test}}$) sets, where the \textbf{atoms} ($\mathcal{A}$) of the domain remain familiar, but the \textbf{compounds} ($\mathcal{C}$)---the specific compositions---are novel. 

We formalize the first condition as $\mathcal{P}(\mathcal{A} | \mathcal{D}_{\text{train}}) = \mathcal{P}(\mathcal{A} | \mathcal{D}_{\text{test}})$, while ensuring the sets of compounds are disjoint:
$$ \texttt{set}(\mathcal{C} | \mathcal{D}_{\text{train}}) \cap \texttt{set}(\mathcal{C} | \mathcal{D}_{\text{test}}) = \emptyset $$
We instantiate this formalization for 3 distinct types of CompGen:

\begin{enumerate}
    \item \textbf{Entity Composition:} This tests generalization to unseen symbols within the familiar logical constraint and rule system.
    \begin{itemize}
        \item \textbf{Atoms ($\mathcal{A}$):} The set of all possible digit symbols, e.g., $\{0, 1, ..., 9\}$.
        \item \textbf{Compounds ($\mathcal{C}$):} The specific subset of digits presented as initial clues in a given puzzle.
        \item \textbf{Test Condition:} $\mathcal{D}_{\text{train}}$ contains puzzles with clues drawn from a subset of digits (e.g., $\{0-4\}$), while $\mathcal{D}_{\text{test}}$ contains puzzles with clues drawn from a disjoint subset (e.g., $\{5-9\}$). The underlying \texttt{all\_different} constraint\footnote{The \texttt{all\_different} ensures that within a specified set of cells, all values must be unique. In Sudoku, this applies to every row, column, and contiguous subgrid.} and rules in $\mathcal{K}_{\text{easy}}$ are constant.
    \end{itemize}

    \item \textbf{Relational Composition:} This tests generalization to novel, fundamental logical rules.
    \begin{itemize}
        \item \textbf{Atoms ($\mathcal{A}$):} The set of all possible fundamental logical axioms. This includes axioms like \texttt{all\_different} and arithmetic axioms (e.g., \texttt{add}, \texttt{multiply}).
        \item \textbf{Compounds ($\mathcal{C}$):} The specific set of axioms that defines a single puzzle's constraints.
        \item \textbf{Test Condition:} $\mathcal{D}_{\text{train}}$ contains puzzles governed only by the \texttt{all\_different} axiom (e.g., standard Sudoku). $\mathcal{D}_{\text{test}}$ introduces novel arithmetic constraints (e.g., involving $\{+, -, \times, \div\}$)
    \end{itemize}

    \item \textbf{Rule Composition:} This tests generalization to novel, multi-step reasoning strategies composed of familiar rules.
    \begin{itemize}
        \item \textbf{Atoms ($\mathcal{A}$):} The set of fundamental logical axioms defined in $\mathcal{K}_{\text{easy}}$.
        \item \textbf{Compounds ($\mathcal{C}$):} The specific chain of deductive steps, or strategy, required to solve a puzzle; axioms defined in $\mathcal{K}_{\text{moderate}}$ and $\mathcal{K}_{\text{hard}}$
        \item \textbf{Test Condition:} $\mathcal{D}_{\text{train}}$ contains puzzles solvable using only simple strategies derivable from $\mathcal{K}_{\text{easy}}$. $\mathcal{D}_{\text{test}}$ contains puzzles that require complex strategies (novel compositions of atoms) represented in $\mathcal{K}_{\text{hard}}$.
    \end{itemize}
\end{enumerate}

Our primary objective is to learn a solver function $F_{\Phi}: \mathcal{Z} \rightarrow \mathcal{Y}$ that minimizes a loss function $\mathcal{L}$ and demonstrates high zero-shot accuracy across these three distinct generalization challenges. This allows us to systematically probe the limits of neuro-symbolic models and analyze the conditions under which different forms of compositional generalization are achieved.

\section{Iterative Logic Tensor Networks ($i$LTN)}\label{sec:method}

Logic Tensor Networks (LTNs) are a neuro-symbolic framework that provides a differentiable bridge between deep learning and first-order logic (FOL) \cite{serafini2016logic,badreddine2022logic}. The core idea is to ground the components of a logical language in a real-valued vector space. Constants are mapped to learnable vector embeddings, and predicates are implemented as neural networks (e.g., MLPs) that output a truth-value in the continuous range $[0, 1]$. By replacing boolean operators with their fuzzy logic counterparts (e.g., using t-norms for conjunction), the satisfiability of any logical formula can be calculated as a differentiable function. This allows a network to be trained via gradient descent to satisfy a complex set of logical axioms. However, standard LTNs are designed for one-shot satisfiability checks and are not naturally suited for sequential, multi-step deduction, which is a key requirement for solving complex reasoning problems.

To address this challenge of complex, multi-step reasoning within a declarative framework, we propose a novel extension to Logic Tensor Networks, which we call \textbf{Iterative Logic Tensor Networks ($i$LTN)}. This methodology enhances the standard LTN framework by introducing an iterative refinement process.

\subsection{The $i$LTN Framework}
The core innovation of the $i$LTN is to decompose a single, complex optimization problem into a sequence of more tractable, iterative steps. Each step involves a localized logical inference that builds upon the deductions of the previous steps. This explicitly models the step-by-step deductive reasoning required for hard combinatorial problems.  By introducing this iterative loop, the $i$LTN is specifically designed to provide a much stronger and more direct mechanism for handling rule composition, as it can now learn to build long chains of reasoning, one step at a time.

\subsubsection{State Representation and Grounding:}
The state of the puzzle at each reasoning step $t \in \{0, \dots, T\}$ is captured by a belief state tensor $P^{(t)} \in ^{N \times N \times |\mathcal{V}|}$. Each entry $P^{(t)}_{r,c,v}$ represents the probability that cell (r, c) holds the symbol $v \in \mathcal{V}$. The initial state $P^{(0)}$ is derived from the visual perception module $G_{\theta}$. All symbolic constants (digits $v \in \mathcal{V}$, cells (r, c)) are grounded as learnable vector embeddings in $\mathbb{R}^{D}$. For our experiments, the embedding dimension is set to $D=64$, and embeddings are initialized randomly from a standard normal distribution ($\mathcal{N}(0, 1)$) before being fine-tuned end-to-end. Logical predicates (e.g., \texttt{is\_digit(cell, digit)}) are implemented as MLPs that operate on these embeddings to produce a truth-value in $[0,1]$.

\subsubsection{The Iterative Refinement Loop:}
The reasoning process unfolds over a variable number of discrete time steps t. At each step:
\begin{itemize}

    \item \textbf{Logical Inference Step:} The model takes its current belief state $P^{(t)}$ and computes a logical loss, $\mathcal{L}_{\text{logic}}^{(t)} = 1 - \text{satisfiability}(\mathcal{K} | P^{(t)})$, based on the axioms in the knowledge base $\mathcal{K}$. The satisfiability of the knowledge base is computed using fuzzy logic operators to ensure differentiability. Specifically, we employ the Lukasiewicz t-norm ($T(a,b) = \max(0, a+b-1)$) for the conjunction operator ($\land$) and a generalized mean operator for the universal quantifier ($\forall$). The model then performs a single gradient step on the belief state $P^{(t)}$ with respect to this loss to produce an updated, more logically consistent belief state, $\tilde{P}^{(t)}$. This step can be viewed as a form of inner-loop optimization, where the belief state tensor itself is momentarily treated as a set of learnable parameters and refined to better satisfy the logical constraints before the main backward pass updates the model's core weights. This can be seen as a form of differentiable belief propagation guided by the logical constraints.

    \item \textbf{Belief State Update with Gumbel-Softmax:} To move from the continuous belief state $\tilde{P}^{(t)}$ to a more discrete state for the next iteration without breaking gradient flow, we use the Gumbel-Softmax relaxation. This allows us to sample a discrete one-hot vector for each cell in a differentiable manner. We employ an annealing schedule for the temperature parameter $\tau$, starting at $\tau = 1.0$ and decaying it exponentially after each training epoch to a minimum of $\tau = 0.1$. This encourages exploration early in training and commitment to discrete hypotheses later on. The result is combined with the previous state to produce the belief state for the next step, $P^{(t+1)}$. This step mimics the human process of committing to a hypothesis and using it for subsequent reasoning.

    \item \textbf{Halting Mechanism:} To handle puzzles of varying difficulty, the model also computes a halting probability $h^{(t)}$ at each step. This probability is computed by a small, two-layer MLP that takes a globally-pooled representation of the current belief state $P^{(t)}$ as input. If $h^{(t)}$ exceeds a threshold, the loop terminates. This allows the model to learn to take as many steps as needed, rather than relying on a fixed horizon $T$.
\end{itemize}

\subsection{Model Training and Objectives}
The model is trained end-to-end to learn predicates that lead to correct deductions.

\subsubsection{Grounding-Only Objective (LTN Baseline):}
This objective trains a model to map an input image $I$ directly to the final puzzle solution $\mathcal{Y}$ in a single step. The goal is to optimize only for the correctness of the final state, without an explicit multi-step reasoning loss. The objective is a standard cross-entropy loss:
\begin{equation}
    \mathcal{L}_{\text{Grounding-Only}} = \text{CrossEntropy}(F_{\Phi}(I), \mathcal{Y})
    \label{eq:grounding_loss}
\end{equation}
Here, $F_{\Phi}$ represents the entire one-shot model.

\subsubsection{Full Objective ($i$LTN):}
The complete end-to-end training objective ensures that at every step $t$, the belief state $P^{(t)}$ moves closer to the ground-truth solution $\mathcal{Y}$. We apply a cross-entropy loss at each iteration, weighted by a discount factor $\gamma$. To accommodate a variable number of steps, the training horizon $T$ is sampled from a distribution for each puzzle. The loss is:
\begin{equation}
    \mathcal{L}_{\text{Full}} = \sum_{t=1}^{T} \gamma^{T-t} \cdot \text{CrossEntropy}(P^{(t)}, \mathcal{Y})
    \label{eq:full_loss}
\end{equation}
This loss supervises the entire reasoning trace. The parameters of the predicates, the perception network, and the halting mechanism are updated via backpropagation through time.
The mathematical form of the loss is identical to the reasoning-only objective. However, unlike the ablated version where the initial state $P^{(0)}$ is derived directly from symbolic data, bypassing the need for perceptual grounding, the initial belief state $P^{(0)}$ is produced by the perception network $G_{\theta}$ operating on the raw input image $I$. 

\subsection{Addressing Compositionality with $i$LTN}
\begin{enumerate}
    \item \textbf{Entity Composition:} The learned predicates are trained to be functions of embeddings. We test if they generalize to operate on novel digit embeddings unseen during training.
    \item \textbf{Relational Composition:} We test if the iterative reasoning process can adapt to a new logical rule. We train using only the \texttt{all\_different} axiom in $\mathcal{K}$. At test time, we add new arithmetic axioms (e.g., for addition or multiplication) to the satisfiability function used in the logical inference step of each iteration.
    \item \textbf{Rule Composition:} This is the key advantage of the $i$LTN. By training on easy puzzles that require few iterations ($T_{\text{easy}}$), we test whether the learned iterative deduction process can scale to hard puzzles that require a significantly larger number of steps ($T_{\text{hard}} > T_{\text{easy}}$) to solve. This directly evaluates the model's ability to compose long chains of reasoning.
\end{enumerate}

\section{Experimental Setup}\label{sec:experiments}
To empirically investigate the relationship between symbol grounding and compositional generalization, we conduct a series of controlled experiments designed to test the hypotheses outlined in Section \ref{sec:problem_statement}. Our setup is structured around a comparative analysis of three models, each trained with a different objective to isolate the contributions of grounding and reasoning.

\subsection{Models and Baselines}
We evaluate three distinct models to isolate the contributions of grounding and reasoning objectives. All models are based on the LTN framework to ensure a controlled comparison.
\begin{enumerate}
    \item \textbf{Grounding-Only LTN:} An LTN architecture trained with a pure grounding objective. The model is tasked with mapping the input image directly to the final puzzle solution, optimizing only for the correctness of the final state using one shot constraint optimization without an explicit reasoning loss.
    \item \textbf{Reasoning-Only $i$LTN (Ablated):} Our $i$LTN model provided with pre-grounded symbolic inputs. This ablative version is trained exclusively on the reasoning objective, isolating the iterative deduction mechanism from the perceptual grounding task.
    \item \textbf{Full $i$LTN:} Our complete end-to-end model, trained jointly on both the visual grounding and the multi-step logical reasoning objectives as described in Section \ref{sec:method}.
\end{enumerate}

\subsection{Dataset and Evaluation Splits}
We generated a dataset of visual logic puzzles, rendered as $84 \times 84$ pixel grayscale images, using ClassicLogic \cite{classic_games_benchmark}. The dataset consists of puzzle images and their corresponding ground-truth solutions. For each of the three CompGen tasks, we generate specific training and test splits to ensure zero-shot evaluation on novel compositions.

Our use of a synthetic visual logic puzzle domain is a deliberate methodological choice. This controlled environment allows us to isolate the relationship between grounding and reasoning from the confounding variable of perceptual uncertainty. By simplifying the perceptual task, we can analyze the emergence of compositional reasoning with a much clearer signal. This aligns with a strong precedent for using synthetic benchmarks to gain precise control to determine whether a model's failure originates from poor grounding or flawed reasoning. 
and rigorously test specific generalization axes.

\subsection{Training Details}
All models are trained for 100 epochs on NVIDIA A100 GPUs. We use the AdamW optimizer with a learning rate of $1 \times 10^{-4}$, $\beta_1=0.9$, $\beta_2=0.999$, and a weight decay of $0.01$. A linear warmup schedule is used for the first 10 epochs, followed by a cosine decay schedule for the remaining epochs. For the $i$LTN model, the iterative reasoning loop is trained with a variable horizon $T \sim \text{Uniform}(5, 20)$. The loss discount factor is $\gamma=0.98$. Within each $i$LTN iteration, we perform $k=3$ gradient steps on the belief state with respect to the logical satisfiability loss, $\mathcal{L}_{\text{logic}}^{(t)} = 1 - \text{satisfiability}(\mathcal{K} | P^{(t)})$, using SGD with a learning rate of $0.1$. The temperature $\tau$ for the Gumbel-Softmax relaxation is annealed linearly from $\tau=1.0$ to $\tau=0.1$ over the first 50 epochs.

\section{Results and Analysis}

Our experiments provide strong empirical evidence for our central hypothesis: while symbol grounding is a necessary prerequisite, it is not sufficient for achieving robust compositional generalization. The full $i$LTN model, which integrates an explicit iterative reasoning mechanism with a grounding objective, consistently and significantly outperforms the Grounding-Only LTN baseline and Reasoning-Only $i$LTN across all three axes of our generalization taxonomy. 

During training, both the Grounding-Only LTN and the Full $i$LTN achieve low symbol grounding loss and high accuracy \textbf{($\approx$95\%)} on the training digits, indicating that both architectures are capable of learning the basic perceptual task of mapping pixels to symbols.


\begin{figure*}[t!]
  \centering
  \includegraphics[width=0.7\linewidth]{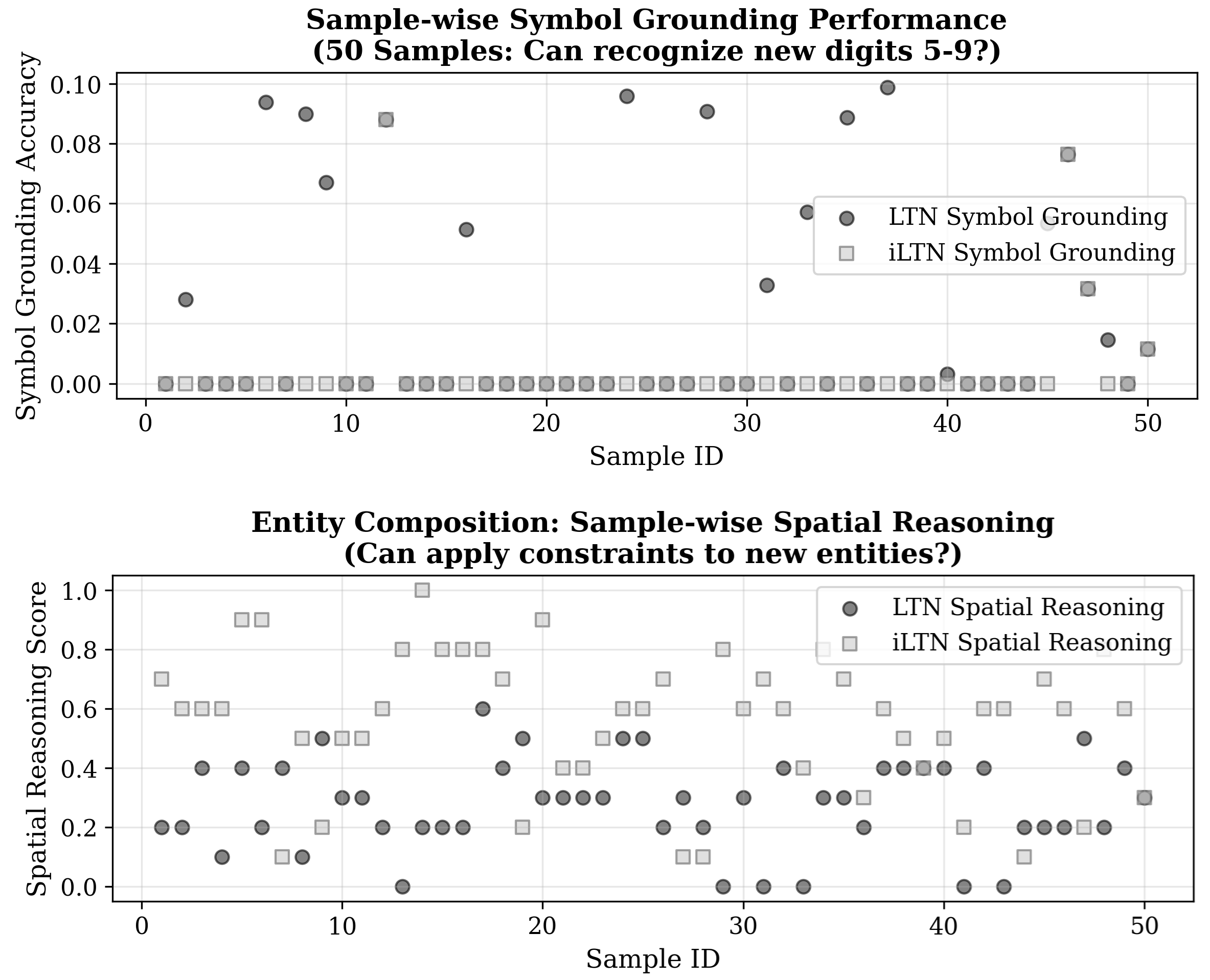}
  \caption{On Entity Composition, both models fail to classify unseen digits, but the $i$LTN is still able to apply logical constraints to them, outperforming the baseline.}
  \label{fig:entity_results}
\end{figure*}

\subsection{Entity Composition: Reasoning Beyond Recognition}
\label{sec:entity_comp}

The Entity Composition task evaluates whether a model can reason using familiar rules over unseen entities (in our case, digits 5--9). As shown in Figure~\ref{fig:entity_results}, both the baseline LTN and our $i$LTN achieve near-zero classification accuracy on these novel digits. This presents an apparent paradox: how can a model reason about constraints between symbols it cannot correctly identify? The answer lies in the distinction between classification and representation. While the final output layer fails to assign the correct labels, the $i$LTN's perceptual module successfully maps the unseen digits to distinct, well-separated clusters in its latent embedding space. 
We visualize this phenomenon in 
using a t-SNE projection of the learned embeddings where we noticed how the cluster for digit '9' (unseen) is proximal to the cluster for '4' (seen), making a classification error plausible. However, the clusters remain separable, allowing a learned logical predicate to distinguish between them. This demonstrates how the model can satisfy logical constraints even when its classification accuracy is low, by leveraging the richer relational structure of the embedding space.
The $i$LTN's logical predicates (e.g., for the \texttt{all\_different} constraint) operate directly on these vector embeddings, not on the final class labels. Because the embeddings for two different unseen digits (e.g., '7' and '8') are metrically distant in the latent space, the predicate can correctly determine they are not the same entity. This allows the $i$LTN to satisfy the puzzle's logical constraints even without accurate classification.

This capability directly translates to a significant performance advantage (Figure.~\ref{fig:all_results}): the $i$LTN solves 31/50 puzzles, whereas the baseline LTN, which is less capable of leveraging the latent structure, solves only 4/50. This result provides strong evidence that robust compositional reasoning relies on learning a structured and separable representation space, rather than on perfect symbol classification alone. 

\subsection{Relational Composition: Adapting to New Rules}
\begin{figure*}[h!]
  \centering
  \includegraphics[width=0.7\linewidth]{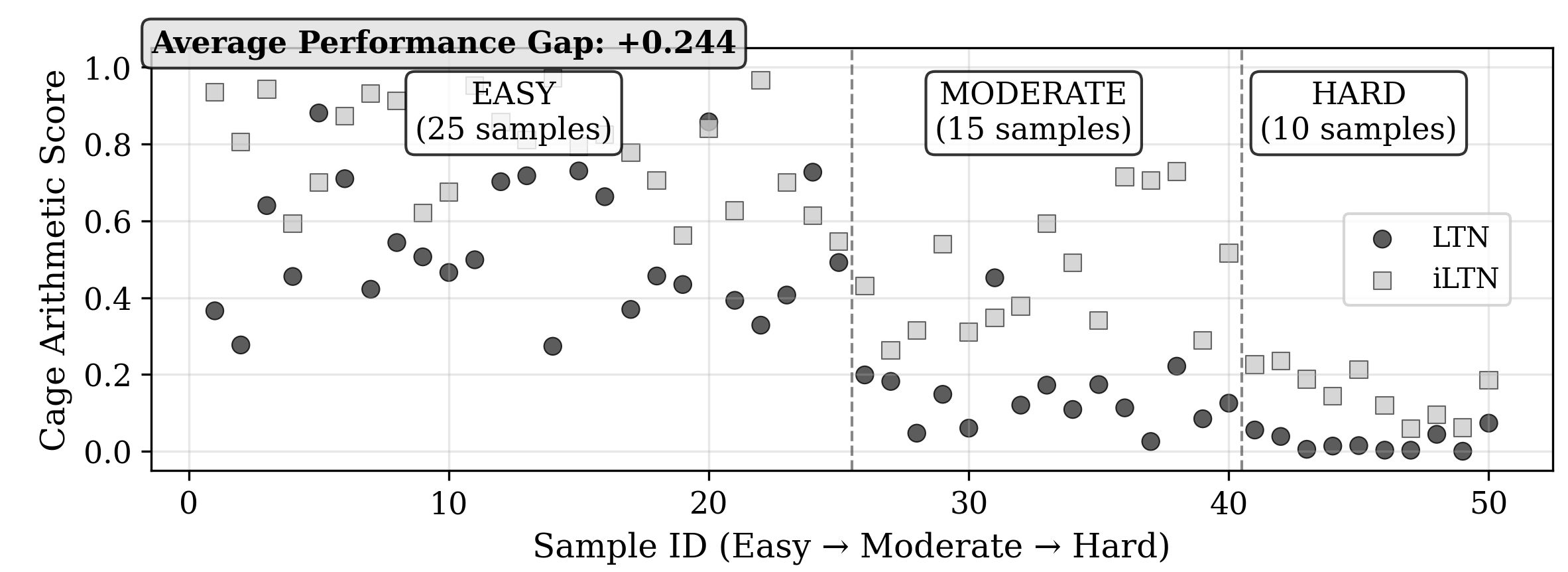}
  \caption{On Relational Composition, $i$LTN demonstrates better generalization by adapting to new arithmetic rules of KenKen.}
  \label{fig:relation}
\end{figure*}

The Relational Composition task evaluates the models' ability to generalize from one set of rules (sudoku's `all\_different' constraint) to a new set that includes novel arithmetic constraints (KenKen). The sample-wise performance plot (Figure~\ref{fig:relation}) shows that the $i$LTN consistently outperforms the baseline, even when the difficulty of the arithmetic constraints increases.

While the baseline model shows some limited success on easy puzzles (22.0\% accuracy), its performance does not scale. The $i$LTN, however, demonstrates a much stronger capacity for transfer, achieving 52.0\% accuracy. This suggests that the iterative reasoning process learned by the $i$LTN is more modular and adaptable, allowing it to incorporate new, unseen logical axioms into its deductive process at test time.

\subsection{Rule Composition: Generalizing to Chain Reasoning}
\begin{figure*}[h!]
  \centering
  \includegraphics[width=0.7\linewidth]{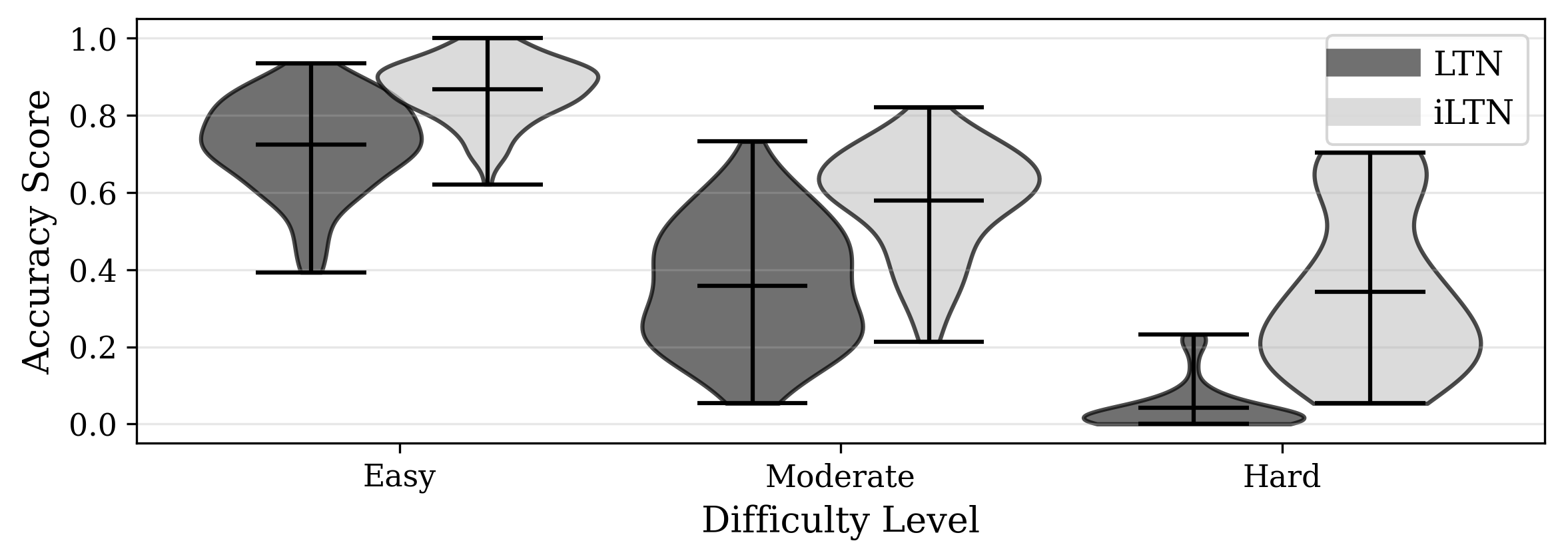}
  \caption{On Rule Composition, unlike the baseline, $i$LTN was able to combine and apply more rules to solve hard strategies.}
  \label{fig:rule}
\end{figure*}

The key architectural advantage of the $i$LTN is its ability to model multi-step deduction. The rule composition task directly tests this by evaluating generalization from easy and moderate puzzles requiring few reasoning steps to hard puzzles requiring long and complex deductive chains. The violin plots in Figure~\ref{fig:rule} illustrate this capability gap perfectly.

On easy puzzles, both models perform reasonably well. However, as puzzle difficulty increases, the baseline LTN's performance collapses, dropping to a mere 4.0\% accuracy on hard puzzles. The $i$LTN, in stark contrast, shows remarkable robustness. Its performance distribution remains high across all difficulty levels, achieving 36.0\% accuracy on the hard set. This demonstrates that the $i$LTN has not simply memorized reasoning paths but has learned a generalizable, iterative deduction policy that can be unrolled for a greater number of steps to solve more complex problems.

\subsection{Synthesis of Results}
\begin{figure*}[h!]
  \centering
  \includegraphics[width=0.7\linewidth]{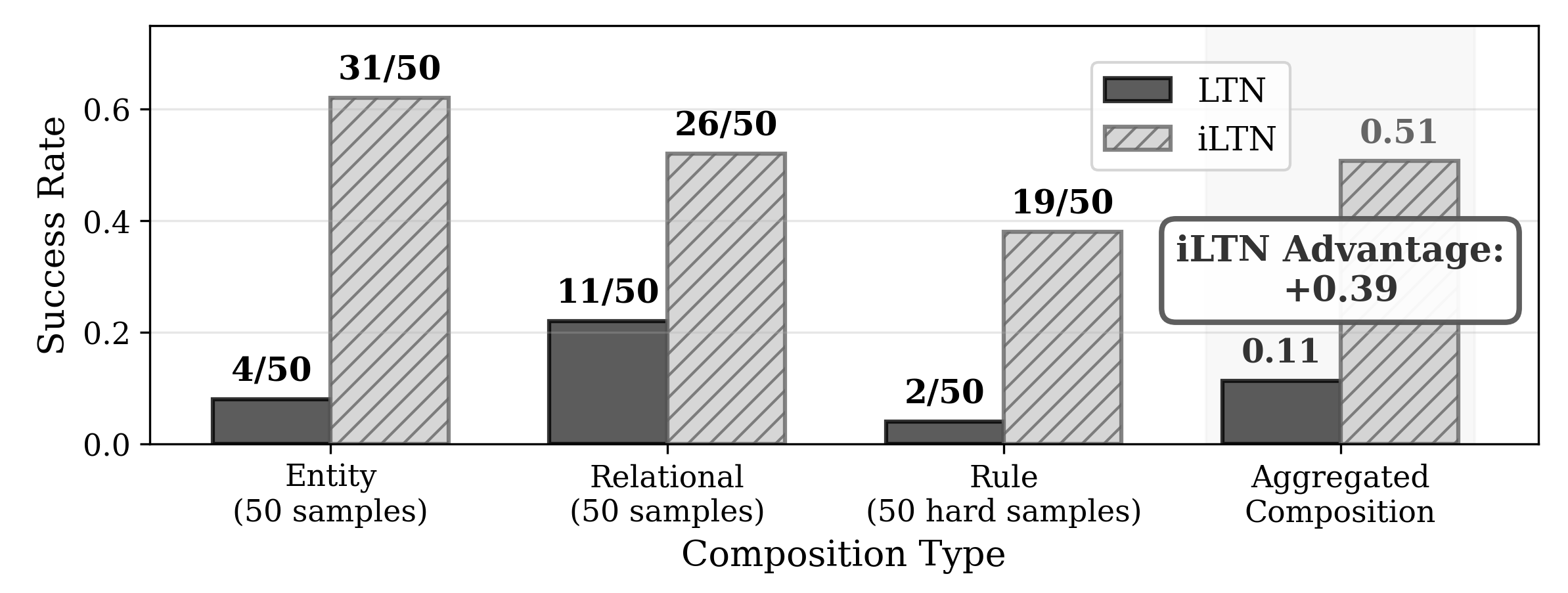}
  \caption{The summary bar visually confirm the $i$LTN's significant and consistent performance across all three axes of compositional generalization.}
  \label{fig:all_results}
\end{figure*}

Figure~\ref{fig:all_results} provides a holistic summary of our findings. It visually confirms that while both models succeed at the foundational task of symbol grounding, the $i$LTN is vastly superior on all other metrics. This directly translates into strong zero-shot performance on all three axes of compositional generalization: entity, relational, and rule composition. The aggregated results cement this conclusion, with the $i$LTN achieving an overall accuracy of 51.2\%—more than four times higher than the 11.3\% achieved by the Grounding-Only baseline. 

\subsection{Quantifying the Interdependence of Grounding and Reasoning}
\begin{figure*}[h!]
    \centering
    \includegraphics[width=1.0\textwidth]{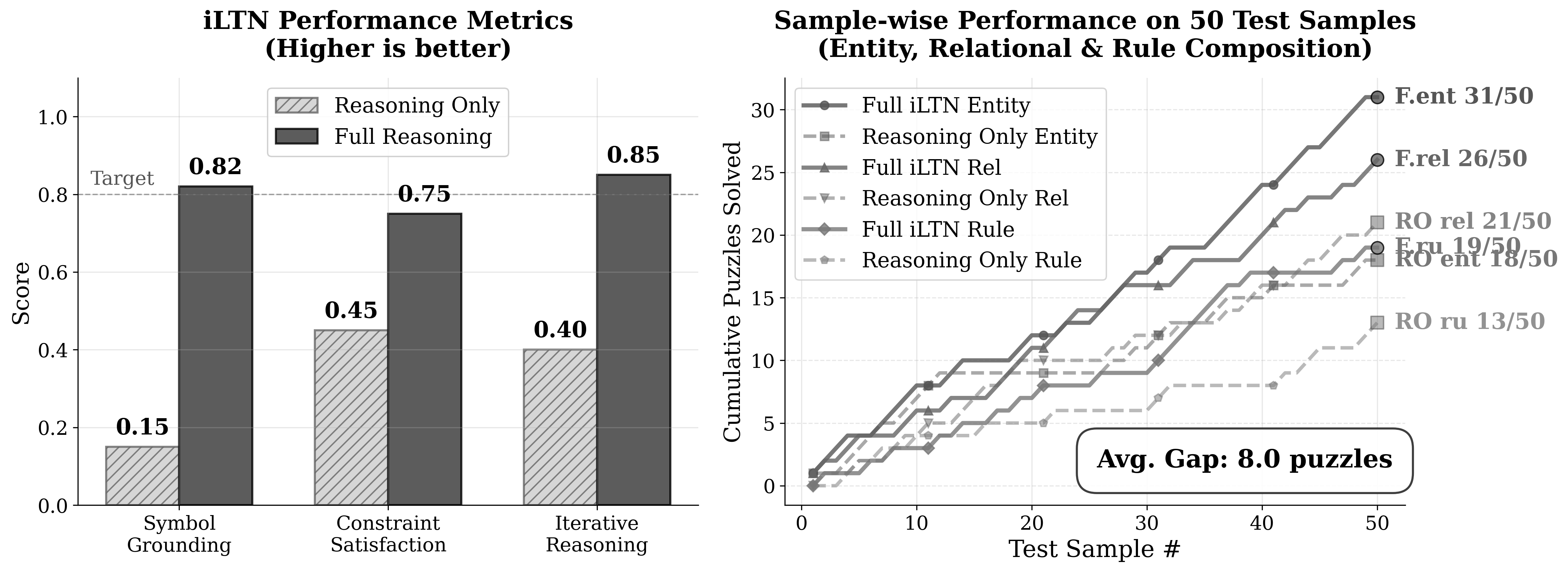}
    \caption{Comparison Performance of Reasoning-Only and Full $i$LTN} 
    \label{fig:main_results_ablation}
\end{figure*}

Figure~\ref{fig:main_results_ablation} (left) presents a direct comparison of core performance metrics between our Full $i$LTN and the Reasoning-Only ablation. The Full $i$LTN, which learns to ground symbols and reason jointly, achieves high scores across all metrics: symbol grounding (0.82), constraint satisfaction (0.76), and iterative reasoning (0.85). Counter-intuitively, the Reasoning-Only model, despite being provided with pre-grounded symbolic inputs, exhibits a drastic performance drop on the reasoning metrics themselves, scoring only 0.45 on constraint satisfaction and 0.40 on iterative reasoning. This finding strongly suggests that the joint, end-to-end training process yields more effective representations for reasoning, where grounding contributes to better generalization capabilities.

This performance differential is further reflected in the downstream compositional generalization tasks, as shown in Figure~\ref{fig:main_results_ablation} (right). The cumulative performance across 50 held-out test samples shows the Full $i$LTN (solid lines) consistently outperforming the Reasoning-Only ablation (dashed lines) across all entity, relational, and rule composition tasks, maintaining an average performance gap of 8.0 solved puzzles. This result directly addresses our central research question with a nuanced conclusion: that symbol grounding may not be sufficient alone for compositionality but plays an important role for reasoning.

\section{Discussion}
Our empirical results provide a clear answer to our central research question: compositional generalization is not an emergent property of symbol grounding. The stark performance gap between the full $i$LTN and the Grounding-Only baseline demonstrates that while grounding is a necessary prerequisite, it is insufficient for enabling robust reasoning.

The baseline's failure highlights that successfully mapping pixels to symbols does not automatically equip a model with the principles governing how those symbols interact if the architecture lacks the capacity for the sequential, multi-step deduction required for complex problems. In contrast, the $i$LTN's success is directly attributable to its iterative refinement loop, which provides an explicit learning objective for compositional reasoning. By training on the entire reasoning trace, the $i$LTN learns a generalizable deduction process, allowing it to compose longer chains of reasoning for difficult, unseen problems.

These findings imply that neuro-symbolic architectures must treat the reasoning process as a first-class citizen, requiring dedicated architectural support beyond a well-trained perception front-end. Future work should test this principle in more diverse domains like robotics or NLP and explore alternative architectures for iterative reasoning.

Furthermore, our ablation study provides a nuanced perspective on the relationship between grounding and reasoning. The counter-intuitive finding that our full $i$LTN outperformed the Reasoning-Only version---which received pre-grounded symbolic inputs—suggests that grounding is more than just a prerequisite. We hypothesize that jointly training the perceptual and reasoning modules acts as a powerful regularizer. By forcing the reasoning module to operate on the noisy and uncertain representations produced by the perception front-end, it learns a more robust and generalizable deductive policy. This suggests that enduring real-world perceptual uncertainty may be a critical element for developing flexible reasoning systems, a promising direction for future neuro-symbolic research.

\subsection{Limitations and Future Work}
Although our findings provide strong evidence for the non-com\-ple\-men\-tar\-ity of grounding and reasoning, we acknowledge certain limitations that open avenues for future research. Our experiments were conducted in a controlled, synthetic visual domain. This was a deliberate methodological choice to isolate the relationship between grounding and reasoning from the confounding variable of perceptual uncertainty in real-world. Future work should test the principles established here in more complex and noisy domains, such as robotics and natural language understanding. Plus, the scalability of the $i$LTN's iterative process to vastly larger problem spaces remains an open question for exploration. 

\section{Conclusion}

This paper tests the assumption that solving symbol grounding is enough for compositional generalization in neuro-symbolic systems and shows that it is not.
Experiments demonstrate that models trained only on grounding fail to generalize to new combinations, even with correct perception. In contrast, the proposed Iterative Logic Tensor Network (iLTN), trained with an explicit multi-step reasoning objective, achieves strong zero-shot performance. Its success comes from iterative refinement that enables complex deductive reasoning.
The contributions are: (1) evidence that grounding and compositional generalization must be optimized together, not separately; (2) a novel architecture that achieves this; and (3) results showing joint training outperforms models trained on grounding or reasoning alone.

\bibliography{aaai2026}

\end{document}